\documentclass[conference]{IEEEtran}
\usepackage{cite}
\usepackage{graphicx}
\usepackage{subfigure}
\usepackage{amsmath} 
\usepackage{amssymb} 
\usepackage{array}

\usepackage{blindtext}
\usepackage{cite}
\usepackage{graphicx}
\usepackage{caption}
\usepackage{tikz}
\usepackage{verbatim} %comentarios
\usepackage{xargs}  
\usepackage[colorinlistoftodos,prependcaption,textsize=tiny]{todonotes}

\usepackage{amsmath}
\usepackage{algorithmic}
\usepackage{xcolor}

\usepackage{array}

\fontfamily{phv}

\def\BibTeX{{\rm B\kern-.05em{\sc i\kern-.025em b}\kern-.08em
    T\kern-.1667em\lower.7ex\hbox{E}\kern-.125emX}}

\usepackage{mwe}
\usepackage{fancyhdr}
\fancypagestyle{firststyle}
{
	\fancyhf[C]{\fontsize{8}{10} \selectfont \textit{2020 IEEE International Autumn Meeting on Power, Electronics and Computing (ROPEC 2020). Ixtapa, Mexico} }
	\fancyfoot[C]{978-1-7281-9953-5/20/\$31.00 \textcopyright 2020 IEEE}
}

\def\histTemp{
	\begin{tikzpicture}

        \node at (5,5) {\includegraphics[width=0.42\textwidth]{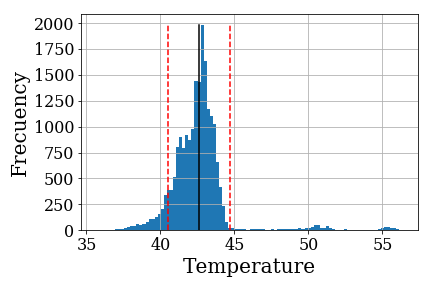}};

        \draw (7.2,4.3)  node[above=5pt,anchor=south,inner sep=0] {\textbf{Hot-spots}};
        
        \draw (6.7,3.7) -- ++(0.5,0.7) node[above=5pt,anchor=south,inner sep=0] {};
        
        \draw (7.9,3.7) -- ++(-0.4,0.7) node[above=5pt,anchor=south,inner sep=0] {};
        
    \end{tikzpicture}
}

\def\guidesc{
\begin{tikzpicture}
  \node[anchor=south west,inner sep=0] (image) at (0,0) {\includegraphics[width=5cm]{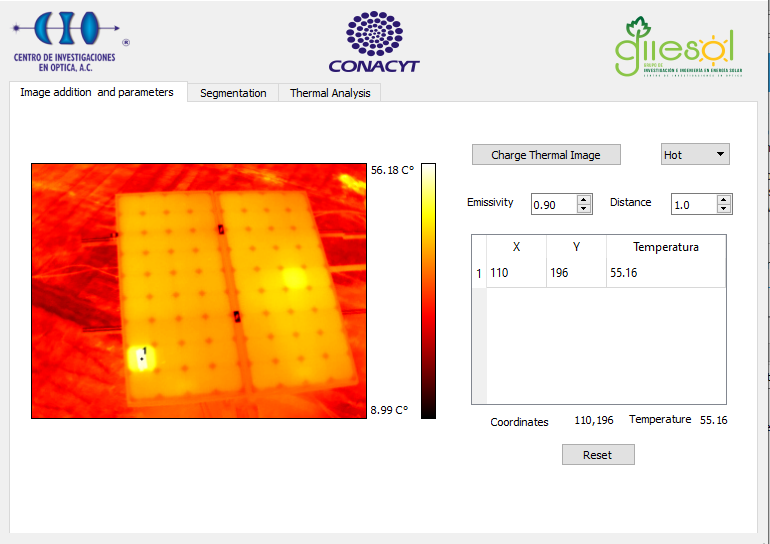}};
  \begin{scope}[x={(image.south east)},y={(image.north west)}]
 
  \draw[<-] (0.7,0.75) -- (0.87,1.1) node at (0.87,1.18) {\small{Load image button}};
  \draw[<-] (0.72,0.15) -- (0.4,-0.04) node at (0.4,-0.1) {\small{Reset button}};
  \draw[<-] (0.95,0.71) -- (1.1,0.7) node[text width=2.5cm] at (1.3,0.7) {\small{Thermal image color map}};
  \draw[<-] (0.85,0.5) -- (1.05,0.2) node[text width=2.5cm] at (1.3,0.2) {\small{Temperature of selected points}};
  \draw[<-] (0.25,0.7) -- (0.08,1.1) node at (0.07,1.18) {\small{Input Image view pane}};

  \end{scope}
\end{tikzpicture}
}

\def\guiseg{
\begin{tikzpicture}
  \node[anchor=south west,inner sep=0] (image) at (0,0) {\includegraphics[width=5cm]{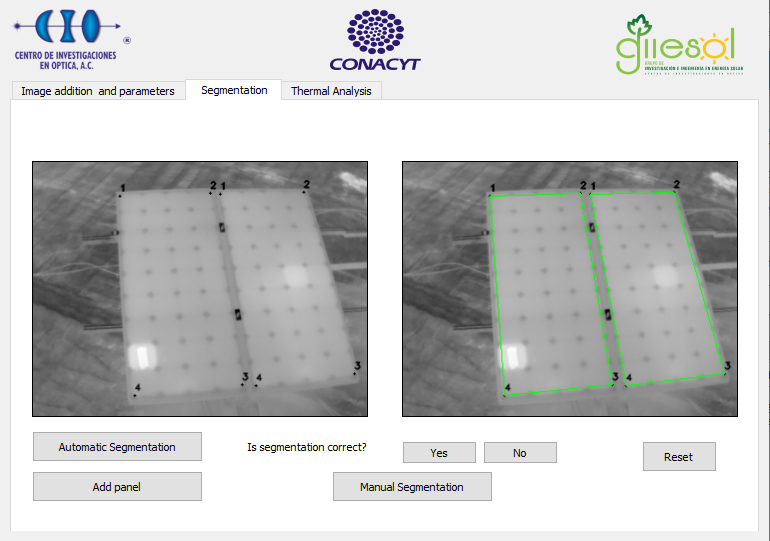}};
  \begin{scope}[x={(image.south east)},y={(image.north west)}]

  \draw[<-] (0.7,0.72) -- (0.87,1.1) node at (0.87,1.18) {\small{PV modules segmented view}};
  \draw[<-] (0.89,0.12) -- (1.03,-0.05) node at (1.05,-0.1) {\small{Reset button}};
  \draw[<-] (0.05,0.06) -- (0,-0.04) node[text width=2.5cm, align=center] at (0,-0.14) {\small{Add manual ROI}};
  \draw[<-] (0.04,0.18) -- (-0.05,0.3) node[text width=2.5cm, align=center] at (-0.2,0.3) {\small{Automatic segmentation}};
  \draw[<-] (0.53,0.03) -- (0.5,-0.04) node[text width=2.5cm, align=center] at (0.5,-0.14) {\small{Manual segmentation}};
  \draw[<-] (0.2,0.72) -- (-0.05,1.1) node at (-0.05,1.18) {\small{Input Image view pane}};

  \end{scope}
\end{tikzpicture}
}

\definecolor{Plum}{RGB}{221,160,221}
\definecolor{OliveGreen}{RGB}{46,139,87}
\newcommandx{\unsure}[2][1=]{\todo[linecolor=red,backgroundcolor=red!25,bordercolor=red,#1]{#2}}
\newcommandx{\change}[2][1=]{\todo[linecolor=blue,backgroundcolor=blue!25,bordercolor=blue,#1]{#2}}
\newcommandx{\info}[2][1=]{\todo[linecolor=OliveGreen,backgroundcolor=OliveGreen!25,bordercolor=OliveGreen,#1]{#2}}
\newcommandx{\improvement}[2][1=]{\todo[linecolor=Plum,backgroundcolor=Plum!25,bordercolor=Plum,#1]{#2}}
\newcommandx{\thiswillnotshow}[2][1=]{\todo[disable,#1]{#2}}

\usepackage{soul}
\usepackage{ulem}

\definecolor{rev_color}{rgb}{0,1,0}
\definecolor{rev_color2}{rgb}{1,0,0}
\setstcolor{red}

\usepackage{ifthen}

\def\markDoc{false}

\ifthenelse{ \equal{\markDoc}{true} }
{
    \newcommand{\edit}[1]{\textcolor{cyan}{#1}}
    
}
{
    \newcommand{\edit}[1]{#1}
    
}
\newcommand{\Real}{{\mathbb{R}}}
\newcommand{\N}{{\mathbb{N}}}

\begin{document}

{\fontfamily{phv}

\title{
Photovoltaic module segmentation and thermal analysis tool from thermal images
}

\author{\IEEEauthorblockN{ Luis E. Montañez }
\IEEEauthorblockA{\textit{Centro de Investigaciones}\\
\textit{en Óptica A.C.}\\
Aguascalientes, Mexico \\
montanezlef@cio.mx}
\and
\IEEEauthorblockN{ Luis M. Valentín-Coronado}
\IEEEauthorblockA{\textit{Centro de Investigaciones} \\
\textit{en Óptica A.C.}\\
Aguascalientes, Mexico  \\
luismvc@cio.mx}
\and
\IEEEauthorblockN{Daniela Moctezuma}
\IEEEauthorblockA{\textit{Centro de Investigaci\'on en Ciencias}\\
\textit{de Informaci\'on Geoespacial A.C.}\\
Aguascalientes, Mexico  \\
dmoctezuma@centrogeo.edu.mx}
\and
\IEEEauthorblockN{ Gerardo Flores }
\IEEEauthorblockA{\textit{Centro de Investigaciones} \\
\textit{en Óptica A.C.}\\
León, Mexico \\
gflores@cio.mx}

}

\maketitle

\thispagestyle{firststyle}
\renewcommand{\headrulewidth}{0in}
\pagestyle{empty}

\pagestyle{fancy}
\chead{\fontsize{8}{10} \selectfont \textit{2020 IEEE International Autumn Meeting on Power, Electronics and Computing (ROPEC 2020). Ixtapa, Mexico} }
\pagenumbering{gobble}

\begin{abstract}
The growing interest in the use of clean energy has led to the construction of increasingly large photovoltaic systems. Consequently, 
monitoring the proper functioning of these systems has become a highly relevant issue. \edit{In this paper, 
automatic detection, and analysis of photovoltaic modules are proposed. To perform the analysis, a module identification step, based on a digital 
image processing algorithm, is first carried out. This algorithm consists of image enhancement (contrast enhancement, noise reduction, etc.), followed by 
segmentation of the photovoltaic module. Subsequently, a statistical analysis based on the temperature values of the segmented module is performed.}
Besides, a graphical user interface has been designed as a potential tool that provides relevant information of the photovoltaic modules.
\end{abstract}

\IEEEpeerreviewmaketitle

\section{Introduction}
Nowadays  more nations believe global climate change is a pressing concern, that is 
why many countries around the world are looking to increase the use of renewable energy and 
consequently decrease the global increase in temperature. Renewable energy, often referred to as clean energy, 
comes from natural sources (i.e. solar energy). Since technological advance has allowed the 
development of increasingly innovative and less-expensive ways to capture and retain clean energy, 
renewables are becoming a very important energy source. Solar energy is one of the most widely used renewable energy sources, 
and the most commonly used solar technology (for homes and businesses) is the solar photovoltaics for electricity.
Although research has focused on the development and improvement of its main element, the photovoltaic (PV) module, it is also necessary to develop tools capable of monitoring 
and evaluating the optimal performance of these main elements.
Monitoring of PV modules is mostly done manually  (which can be extremely
time-consuming), through the visual inspection or by means 
of the I-V\edit{(current-voltage characteristic)} curve measurement and analysis\edit{\cite{Aranda09}}. 
In addition to visual inspection, another commonly used approach
to fault detection is by detecting hot-spots in the module from a thermographic analysis.

Hot-spots occur when at least one solar cell is damaged, causing it to begin dissipating energy, in the form of heat, rather than producing electrical energy.  Commonly a module will present a hot-spot if a cell is mechanically damaged, there is damage to the bypass diode or failure of the contact points~\cite{belleza1}.

The infrared (IR) thermography technology has gained more recognition and acceptance due to its non-contact 
and non-destructive inspection. It is a fast and  operates without interrupting the operation of
power system.
 
The fault diagnosis, in IR thermograph based technique, is performed through the analysis of the image captured by an infrared camera~\cite{buerhop2012reliability}.

In this work, an automatic photovoltaic module segmentation from thermal images, acquired by a thermal camera mounted on a air vehicle,
as well as the statistical thermal analysis of the segmented module are proposed. 
In addition, the proposed methodology has been implemented on a graphical user interface, in order to provide a practical and useful analysis tool.

The remainder of this work is organized as follows. In 
Section~\ref{sec:related_workd} the related work is presented. 
Section~\ref{sec:methodology} presents the proposed methodology for 
the segmentation and thermal analysis of photovoltaic modules.
In Section~\ref{sec:results} the results that were obtained are shown.
Finally, Section~\ref{sec:conclusions} presents the conclusions as 
well as the future work.

\section{Related work}\label{sec:related_workd}
IR thermography  is a reliable and precise tool for the diagnosis of defects, both optical and electrical, in photovoltaic cells and, 
in addition, for the identification of the precise location and severity of these failures\cite{kaplani2012detection}. This diagnosis is made through analysis of thermal images that could be acquired from ground or from a camera mounted on an air vehicle. Ground thermal images could be used 
for detecting defective modules but the main problem with this technique is that it takes plenty of time. For example, to inspect a three megawatts (MW) phtovoltaic power plant
with thermal images acquired from the ground, a time period of around 34 days is required, while performing the same inspection using thermal air images will take approximately 3 hours~\cite{gallardo2018image}. 
Thus, aerial visual inspection has proven to be a good
option since measurements and fault detection (and even its classification), can be done in short
time periods and with good results~\cite{belleza1}.
Taking advantage of the potential of thermal air image analysis, works such as the one presented in~\cite{tsanakas2015fault}, have proposed a strategy for the generation of an orthomosaic and from there carry out the detection of faults of an entire photovoltaic system. On the other hand, combining the use of a thermal camera and a standard digital camera has made it possible to extend the set of faults that can be detected, as it shown in the work presented in~\cite{grimaccia2017survey}.

Nevertheless, regardless of whether the images come from a source on the ground or in the air, decoding and analyzing the thermal image information
manually, will take a considerable amount of time. Then, to address this issue some of the state of the art works have focused on the
automatic detection of faulty modules in thermal images. For instance, in the work presented in~\cite{tsanakas2015fault2}, the authors have proposed 
an automatic extraction of the so-called ROI (region of interest) using the Canny edge detector algorithm to then perform the fault detection on the PV modules. 
Similarly, in the work presented in~\cite{aghaei2016pv}, the authors have used the Sobel edge detector operator to generate an image mosaicing to be able of analyzing it to detect defects and healthy modules in PV strings.
Even though the PV modules detection in an image can be done using Canny's algorithm or Sobel operator, in most cases the presence of artifacts considered as noise, in the resulting image, is still a problem, that is why in the work presented in~\cite{kim2016automaticarea} the authors have proposed the use of morphological operations to eliminate these artifacts.

Another approach to detect defects in photovoltaic modules is the one proposed in~\cite{jaffery2017scheme}, which is based on fuzzy rules.
These fuzzy rules take into account the similarities, by comparing pixel by pixel, a good and faulty modules images as well as their temperature values.

Besides of this soft computing approach, some authors have used a statistical approach to detect defects in a photovoltaic 
modules~\cite{kim2016automatic,aghaei2015innovative,francesco2018semi}.
Another interesting approach for the detection of failures in photovoltaic modules is the use of Deep Learning as those proposed in \cite{Xie_2020,Pierdicca_2018}.

\section{Methodology}\label{sec:methodology}

A photovoltaic system can be analyzed through the inspection of each of its elements (modules). 
Defect detection in the modules could be performed by infrared thermography, which detects infrared 
energy emitted from object, converts it to temperature, and displays thermal image (thermogram) 
of temperature distribution. 
\begin{figure}[h]
	\centering
	\includegraphics[width=0.25\textwidth]{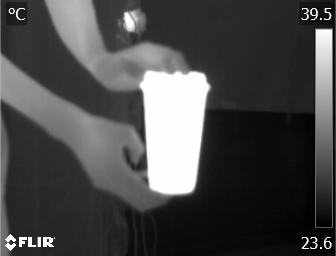}
	\caption{Digital representation of a thermal image.}
	\label{fig:thermal_img}
\end{figure}
A thermal image is a digital representation of a scene and a measure of the thermal radiation 
emitted by the pictured objects. With a thermal image, it is possible to remotely sense the 
temperature of an object or at least accurately indicate its temperature relative to its environment. 
In Fig.~\ref{fig:thermal_img} an example of a digital representation 
of a thermal image is shown.

Given that a thermal image has a digital representation, then, digital image processing techniques 
can be used to perform the module segmentation and thus carry out the detection of failures 
from a statistical analysis of the temperature.

The proposed methodology for detecting defects in photovoltaic modules consists of two main stages, as described in Fig.~\ref{fig:methodology}.
In the first one, the photovoltaic module is extracted using digital image processing algorithms detailed in Section~\ref{sec:pvm_seg}. While, 
in the second stage, the segmented module is used to proceed with the thermal analysis in order to detect defects in the module, as it is explained in Section~\ref{sec:pvm_thermal_analysis}.
\begin{figure}[h!]
	\centering
	\includegraphics[scale=0.42]{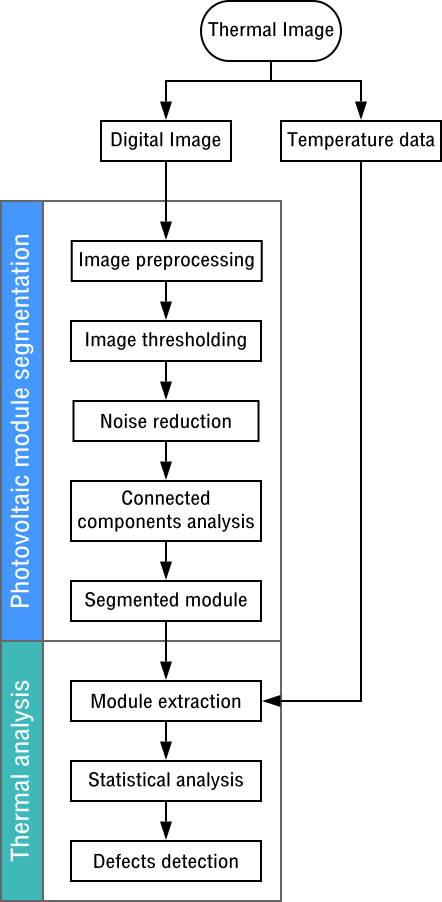}
	\caption{Flow chart of the proposed methodology.}
	\label{fig:methodology}
\end{figure}
\subsection{Photovoltaic module segmentation.}
\label{sec:pvm_seg}

As a first step in detecting defects in the photovoltaic module, it is needed to segment it. To achieve this, the following 
procedure is proposed.

Let $I$, the visual image of the thermogram, a function,
\begin{equation}\label{eq:image}
	I:\mathcal{U}\to [0,1]^c
\end{equation}
where $\mathcal{U}=\{0,\dots,m-1\}\times \{0,\dots,n-1\}$ are the pixels, 
$m\in \N$ and $n\in \N$ the number of rows and columns, 
and $c\in \{1,3 \}$ 
is the number of channels. $I(i,j,c)$ would then give the intensity of the image at pixel position $(i,j,c)$. In particular
if $c=3$, then a color image, in the RGB space, can be denoted as $I(i,j,c)=(\,I_r(i,j),\,I_g(i,j),\,I_b(i,j)\,)$, where $I_r(i,j)=I(i,j,1),\, I_g(i,j)=I(i,j,2)$ and $I_b(i,j)=I(i,j,3)$ refer to the Red, Green and Blue channels respectively.\\

\textbf{Image preprocessing.}\\
To segment the photovoltaic module a color space change is performed, transforming the image $I$ 
into a grayscale image, denoted by $G$; this conversion is given by $G(i,j) = 0.299 I_r(i,j,1) + 0.587 I_g(i,j,2)+ 0.114 I_b(i,j,3)$ \cite{lumareference}. 

Usually, the visual image of the thermogram typically has low contrast, then, implementing some method that allows improving the contrast is convenient.
\begin{figure}[h]
	\centering
	\subfigure[Input image.] {
		\includegraphics[width=0.22\textwidth]{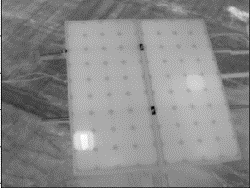}
		\label{fig:input}
	}
	\subfigure[Original histogram.] {
		\includegraphics[width=0.22\textwidth]{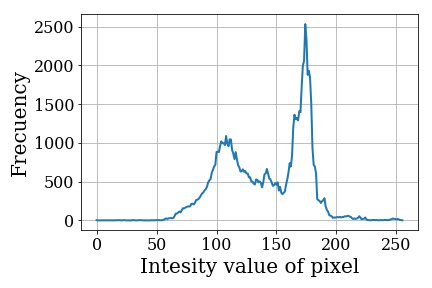}
		\label{fig:hist}
	}
	\subfigure[Image with enhanced contrast and Gaussian noise reduction.] {
		\includegraphics[width=0.22\textwidth]{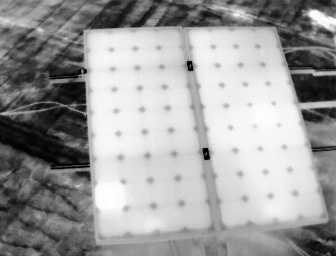}
		\label{fig:dhe_clahe}
	}
	\subfigure[Improved image histogram] {
		\includegraphics[width=0.22\textwidth]{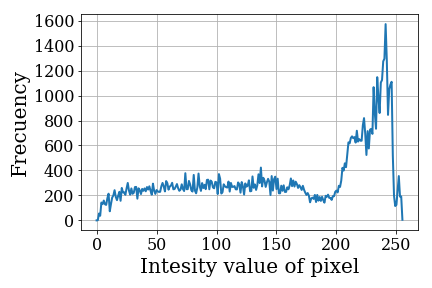}
		\label{fig:hist_dhe_clahe}
	}
	\caption{Image contrast improvement through the Dynamic Histogram Equalization and the Equalization of Adaptive Limited Contrast Histogram techniques.}
	\label{fig:histogram_equalization}
\end{figure}
Although a global histogram equalization approach 
is suitable for contrast improvement, it is not adapted to the local brightness characteristics of the thermal image  equalization~\cite{reza2004realization}.
To avoid this problem a Dynamic Histogram Equalization (DHE) followed by the application of the Equalization of Adaptive 
Limited Contrast Histogram (\textit{CLAHE}) technique were implemented. DHE decomposes the global image histogram into a 
number of sub-histograms based on their local minima and assigns specific gray level ranges for each partition before equalizing them separately~\cite{abdullah2007dhe}, while CLAHE technique works on very small regions in the image, so that only the 
set of pixels that falls in this region is taken into account. Besides, 
once the contrast has been improved, a Gaussian noise reduction is performed. 
In Fig.~\ref{fig:histogram_equalization} the resulting image (which will be denoted as $G^*(i,j)$) is shown. \\

\textbf{PV Module segmentation.}\\

To separate the photovoltaic module of the background a thresholding technique is used. One way to do that is by selecting a fixed threshold 
value $th$, so that  the new pixel value  will depend on this threshold, as follows:

\begin{equation*}
B(i,j)
=\left\{
\begin{array}{ll}
0 \quad \text{if } \,\, G^*(i,j) < th\\
1 \quad \text{if } \,\, G^*(i,j) \geq th
\end{array}
\right.
\end{equation*}
where $B(i,j)$ is the resulting binary image of this process

However, the choice of the threshold value is not trivial, some thresholding methods select this value based on  image characteristics 
or the desired result. Since the characteristics of a thermal image change, it is not practical to establish a fixed value for this threshold.
An adaptive thresholding algorithm that calculates different thresholds for different regions of the same image will have better results. To 
perform an automatic image thresholding, the  Otsu's method has been implemented~\cite{otsu1979threshold}.
 
Subsequently, by using the opening morphological operation~\cite{Burgeth2004}, which is defined as
\begin{equation}
B \circ K = (B\ominus K)\oplus K
\end{equation}
where $B$ is the previously obtained binary image,
$K\in\Real^2$ is the so-called structuring element that determines the neighborhood relation of pixels with respect to a 
shape analysis task, $\ominus$ is an erotion operation determined by,
\begin{equation*}
	(B\ominus K)(i,j) = \inf\{B(x+x',y+y')|(x',y')\in K\}
\end{equation*}
and $\oplus$ is a dilatation operation defined as,
\begin{equation*}
(B\oplus K)(i,j) = \sup\{B(x-x',y-y')|(x',y')\in K\}
\end{equation*}
the area of the photovoltaic modules are segmented, besides of remove the imperfections in the structure of image.

\begin{figure}[h!t!b]
	\centering
	\subfigure[Image thresholding using the Otsu's method] {
		\includegraphics[width=0.22\textwidth]{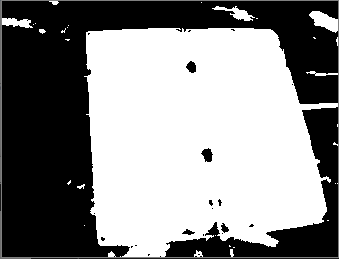}
		\label{fig:otsu}
	}
	\subfigure[Morphological operation] {
		\includegraphics[width=0.22\textwidth]{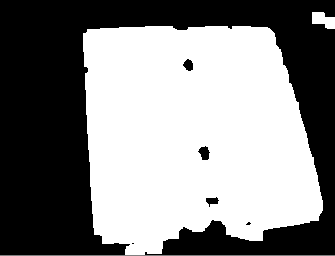}
		\label{fig:opening}
	}
	\caption{Photovoltaic module segmentation. Fig.~\ref{fig:otsu} shows the image thresholding using the Otsu's method, while Fig.~\ref{fig:opening} shows the image once the morphological operations have been applied}
	\label{fig:panel_seg_1}
\end{figure}

In Fig.~\ref{fig:opening} the result of the photovoltaic modules segmentation process is shown. Even though the segmentation
result provide a good estimation of the photovoltaic module in the image, there are some regions where there is still part of the
background. To refine modules segmentation the watershed transform is used. Nevertheless, watershed transform works better if 
a set of ``markers'' are used \cite{roerdink2000watershed}. A marker is a set of pixels that belongs to image foreground objects.
In order to define these markers the distance transform and a connected components approach  have been implemented.

In Fig.~\ref{fig:panel_seg_2} the segmentation of the PV modules, after using the watershed segmentation algorithm, is shown. The resulting image, which will be denotes as $I_{PV}$, will be used for thermal analysis.

\begin{figure}[h!t!b]
	\centering
	\subfigure[PV modules segmentation (binary image)] {
		\includegraphics[width=0.22\textwidth]{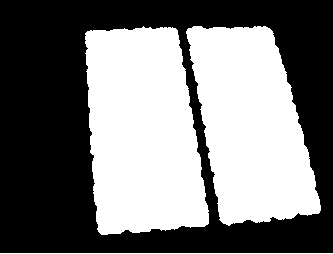}
		\label{fig:binary_modules}
	}
	\subfigure[Border of the segmented modules marked in the input visual thermal image.] {
		\includegraphics[width=0.22\textwidth]{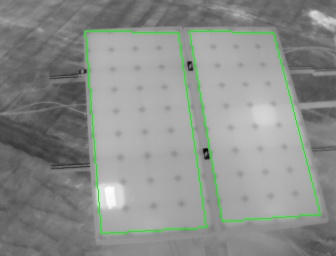}
		
	}
	\caption{Photovoltaic modules segmentation result.}
	\label{fig:panel_seg_2}
\end{figure}

\subsection{Analysis of photovoltaic modules}
\label{sec:pvm_thermal_analysis}

As mentioned, a thermogram contains both visual information as well as temperature information of the area captured by the thermal camera.
Both information can be represented by matrix arrangements of the same dimensions.

Let $T$, the temperature information of the thermogram, a function,
\begin{equation}
T:\mathcal{U}\to [0,1]
\end{equation}
where 

$\mathcal{U}=\{0,\dots,m-1\}\times \{0,\dots,n-1\}$
indicate the position on the temperature matrix (which correspond one-to-one with the position of 
the pixel in image $I$) and $m\in \N$ and $n\in \N$ the number of rows and columns. $T(i,j)$ would then give the temperature value position $(i,j)$. 
Since $T$ and $I$ are the same size, then it is possible to use the module segmentation, performed on image $I$, to carry out a the temperature analysis. Let $T_{PV}\subset T$, such that $T_{PV}(k,l)$ would give the module temperature value 
at position $(k,l)$, determined by,
\begin{equation}
\begin{split}
	T_{PV}(k,l):=\{ T(i,j)\,|&\,I_{PV}(i,j)=1,\\
	&\forall\;i\in [0,m-1],\,j  \in[0,n-1]\}
\end{split}
\end{equation}
where $I_{PV}$ corresponds to the binary image of the segmented photovoltaic modules (see Fig.~\ref{fig:binary_modules}).

\subsection{Defects detection}
Typically a photovoltaic module operating under normal conditions has a uniform temperature distribution, however, if there is a defect 
within the module, a significant change of the temperature occurs \cite{francesco2018semi}.
To observe these temperature variations throughout 
the module, i.e. temperature distribution, a
histogram has been used. Also, from this histogram,
it is possible to define a reference temperature
threshold that allows identifying the defects
in the module. The selection of a fixed 
temperature threshold is not trivial, for instance,
in the work presented in~\cite{tsanakas2013detection} the authors
consider a temperature difference of $5^{\circ}$C 
between the expected temperature and measured 
temperature, while in~\cite{moreton2014dealing}, 
the authors recommend a temperature difference 
of $10^{\circ}$C. Thus, using a fixed temperature
value to determine if a defect exists is not the
best alternative. A better strategy to determine
the temperature value from which a defect is 
considered to exist in the module is from a 
statistical analysis, as shown in the work 
presented in~\cite{kim2016automatic}. In this work, a similar approach that the one presented in~\cite{kim2016automatic} has been implemented.
 As a firs step of this analysis a the mean temperature ($\mu_t$), from the $T_{PV}$ matrix, is calculated as follows,
\begin{equation}
	\mu_t =\frac{1}{n} \left(\sum_k\sum_l T_{PV}(k,l)\right)
\end{equation}
as well as the standard deviation ($\sigma_t$), which is defined as,
\begin{equation}
	\sigma_t = \sqrt{\frac{1}{n}\sum_k\sum_l \Big(T_{PV}(k,l) - \mu_t\Big)}
\end{equation}
From these values and the histogram, it is possible to determine that a module defect can be identified from a temperature greater than one standard deviation away from the mean, as it is shown in Fig.~\ref{fig:histogram_temperature}.
\begin{figure}[h!t!b]
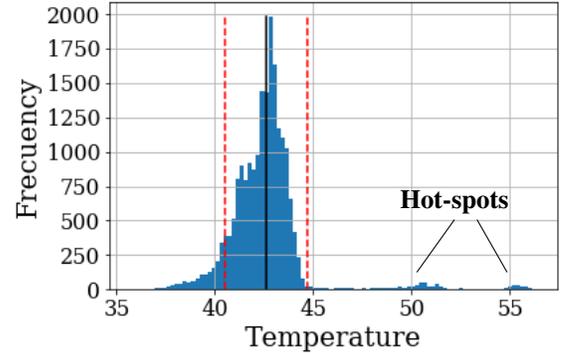

    \centering
        \histTemp
    \caption{Histogram showing the temperature distribution. The black line represents the mean, while the red line represents one standard deviation away from the mean.}
    \label{fig:histogram_temperature}
\end{figure}

Then, to identify defects on the module a temperature threshold $th_t = \mu_t+\sigma_t$, can be used, as follows,
\begin{equation}
    F = \{(k,l)\,|\,T_{PV}(k,l)>th_t\}
\end{equation}
where $F$ is the set of coordinates of the module in which a defect is presented.
\begin{figure}
	\centering
	\includegraphics[width=0.25\textwidth]{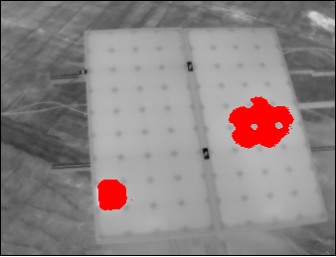}
	\caption{Defects detected in each module.}
	\label{fig:defect}
\end{figure}

Finally, in Fig.~\ref{fig:defect} the result of the defects detection process on each segmented module is shown. It is worth to mention that each of the modules has been analyzed independently, that is, for each module, a statistical analysis has been carried out as described above.

\section{Results}\label{sec:results}

The proposed approach has been tested on sample thermal  images with different conditions from \cite{alfaro2019dataset}, acquired by a thermal camera (Zenmuse XT IR camera) mounted on an unmanned aerial vehicle (DJI Matrice 100).

Fig.~\ref{fig:thermal_img_r1_orig} shows a thermal image that does not appear to have any defects, sometimes, this phenomenon may occur, however after carrying out the segmentation of the modules and their subsequent statistical thermal analysis, it can be seen, as the Fig.~\ref{fig:thermal_img_r1_def} shows, that in fact, there are hot spots that represent module defects.

\begin{figure}[h!t!b]
    \centering
    \subfigure[Thermal image without apparent defects.] {
		\includegraphics[width=0.22\textwidth]{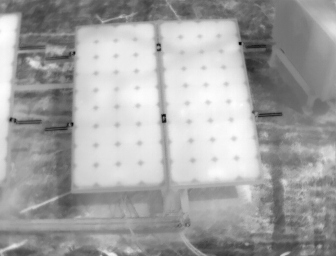}
		\label{fig:thermal_img_r1_orig}
	}
	\subfigure[Defects detected in thermal image. ] {
		\includegraphics[width=0.21\textwidth]{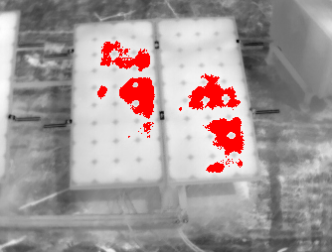}
        \label{fig:thermal_img_r1_def}
	}
    \caption{Analysis of thermal image without apparent defects. }
    \label{fig:thermal_img_r1}
\end{figure}

Another possible scenario is in which a PV module may partially appear in the thermal image, as it is shown in Fig.~\ref{fig:thermal_img_r2_orig}. Nevertheless, the proposed methodology is capable of dealing with this scenario. In the Fig.~\ref{fig:thermal_img_r2_def}, the detection of defects from the proposed approach is shown.
\begin{figure}[h!t!b]
    \centering
    \subfigure[Thermal image with complete  and partial PV modules.] {
		\includegraphics[width=0.22\textwidth]{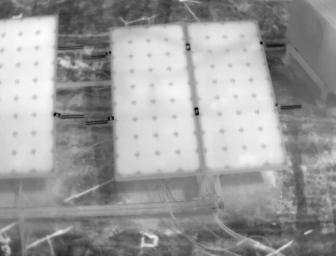}
		\label{fig:thermal_img_r2_orig}
	}
	\subfigure[Defects detected in thermal image with complete  and partial PV modules. ] {
		\includegraphics[width=0.22\textwidth]{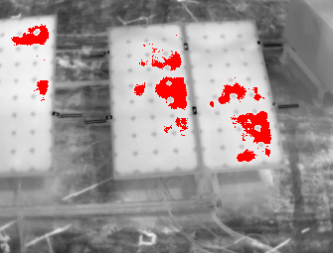}
        \label{fig:thermal_img_r2_def}
	}
    \caption{Analysis of thermal with partially present PV module.}
    \label{fig:thermal_img_r2}
\end{figure}

\edit{As can be seen, the proposed methodology has been tested under different conditions achieving good results, nevertheless, it would be interesting to test the method using images acquired with controlled and well-known parameters, both for the capture and the status of the PV modules (known defects and/or modules without any failure).}

Certainly, automatic detection of defects in photovoltaic modules based on thermal imaging is a useful and practical tool. 
 For this reason, a graphical user interface (GUI) capable of showing, segmenting and subsequently thermally evaluating the present module or modules
 (according to the previously proposed methodology), has been implemented. In the proposed GUI, the user can select a point on the image and obtain the temperature value as a result. In the Fig.~\ref{fig:gui} the designed GUI is shown.
 \begin{figure}[htbp]
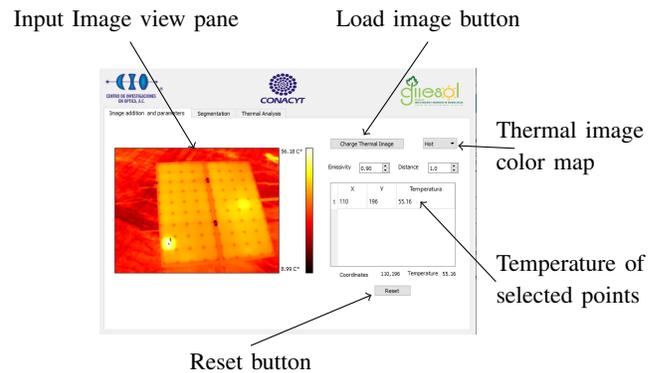

 	\centering
    \guidesc
 	\caption{Graphical user interface. On the left side, the thermal image is shown, besides, temperature information is displayed if the user clicks on the image, showing the temperature in the table of the left side.
 	}
 	\label{fig:gui}
 \end{figure}

Once the thermal image has been loaded, the modules segmentation can be carried out through the option implemented in the GUI, as it is shown in Fig.~\ref{fig:gui_seg}.

\begin{figure}[htbp]
	\centering
	\guiseg
	\caption{Segmentation options implemented on the GUI. On the left side, the loaded image is shown, while on the right side, the segmentation of each of the modules can be seen.}
	\label{fig:gui_seg}
\end{figure}

\begin{figure}[htp]
	\centering
	\includegraphics[width=0.45\textwidth]{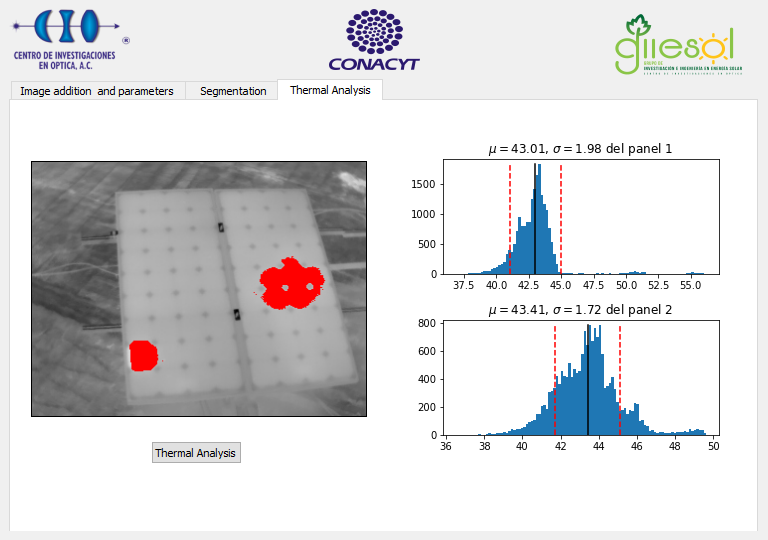}
	\caption{GUI thermal analysis option.
	On the left side, the detected defects are marked in red. On the right side, the temperature histogram of each panel is shown.}
	\label{fig:gui_analysis}
\end{figure}

Finally, as the last option of the GUI, the defects detection by means of the statistical thermal analysis, has been implemented. As it is shown in
Fig.~\ref{fig:gui_analysis}, the detected defects as well as the module temperature distribution can be appreciated.
\section{Conclusions}\label{sec:conclusions}
In this work, a segmentation methodology of PV modules, based on image processing techniques, from thermal images as well as 
a statistical thermal analysis have been proposed. The statistical thermal analysis showed that cells of the PV module without defects will have temperature values within the range of one standard deviation away from the mean, while cells with defects will have higher temperature values. Thus, the proposal of a variable threshold based on the thermal distribution of each module helps to identify the defective ones.
Furthermore, a GUI has been designed, which can be considered as  
a potential tool for thermographic analysis of photovoltaic modules
as it provides relevant information on the state of the modules, 
as well as being a non-destructive and non-invasive inspection approach. 
As future work, deep learning techniques to segment and identify defects on the PV modules will be tested.

\section*{Acknowledgment}
Thanks to CONACYT for the financial support of the scolarship 957872 of "BECAS NACIONALES" .
\bibliography{bibliography.bib}

\begin{thebibliography}{10}

\bibitem{abdullah2007dhe}
M.~Abdullah-Al-Wadud, M.H. Kabir, M.A.A. Dewan, and O.~Chae.
\newblock A dynamic histogram equalization for image contrast enhancement.
\newblock {\em IEEE Transactions on Consumer Electronics}, 53(2):593--600,
  2007.

\bibitem{aghaei2015innovative}
M.~Aghaei, F.~Grimaccia, C.A. Gonano, and S.~Leva.
\newblock Innovative automated control system for pv fields inspection and
  remote control.
\newblock {\em IEEE Transactions on Industrial Electronics}, 62(11):7287--7296,
  2015.

\bibitem{aghaei2016pv}
M.~Aghaei, S.~Leva, and F.~Grimaccia.
\newblock Pv power plant inspection by image mosaicing techniques for ir
  real-time images.
\newblock In {\em Photovoltaic Specialists Conference}, pages 3100--3105. IEEE,
  2016.

\bibitem{alfaro2019dataset}
E.~Alfaro-Mej{\'\i}a, H.~Loaiza-Correa, E.~Franco-Mej{\'\i}a, and S.E.
  Nope-Rodr{\'\i}guez A.D. Restrepo-Gir{\'o}n.
\newblock Dataset for recognition of snail trails and hot spot failures in
  monocrystalline si solar panels.
\newblock {\em Data in brief}, 26:104441, 2019.

\bibitem{Aranda09}
E.~D. {Aranda}, J.~A. {Gomez Galan}, M.~S. {de Cardona}, and J.~M. {Andujar
  Marquez}.
\newblock Measuring the i-v curve of pv generators.
\newblock {\em IEEE Industrial Electronics Magazine}, 3(3):4--14, 2009.

\bibitem{buerhop2012reliability}
Cl. Buerhop, D.~Schlegel, M.~Niess, C.~Vodermayer, R.~Wei{\ss}mann, and CJ.
  Brabec.
\newblock Reliability of ir-imaging of pv-plants under operating conditions.
\newblock {\em Solar Energy Materials and Solar Cells}, 107:154--164, 2012.

\bibitem{Burgeth2004}
B.~Burgeth, M.~Welk, C.~Feddern, and J.~Weickert.
\newblock Morphological operations on matrix-valued images.
\newblock {\em Computer Vision - ECCV 2004. Lecture Notes in Computer Science},
  3024(1):155--167, 2004.

\bibitem{gallardo2018image}
S.~Gallardo-Saavedra, L.~Hern{\'a}ndez-Callejo, and O.~Duque-Perez.
\newblock Image resolution influence in aerial thermographic inspections of
  photovoltaic plants.
\newblock {\em IEEE Transactions on Industrial Informatics}, 14(12):5678--5686,
  2018.

\bibitem{grimaccia2017survey}
F.~Grimaccia, S.~Leva, A.Dolara, and M.~Aghaei.
\newblock Survey on pv modules’ common faults after an o\&m flight extensive
  campaign over different plants in italy.
\newblock {\em IEEE Journal of Photovoltaics}, 7(3):810--816, 2017.

\bibitem{francesco2018semi}
F.~Grimaccia, S.~Leva, and A.~Niccolai.
\newblock A semi-automated method for defect identification in large
  photovoltaic power plants using unmanned aerial vehicles.
\newblock In {\em Power \& Energy Society General Meeting (PESGM)}, pages 1--5.
  IEEE, 2018.

\bibitem{jaffery2017scheme}
Z.A. Jaffery, A.K. Dubey, A.~Haque, et~al.
\newblock Scheme for predictive fault diagnosis in photo-voltaic modules using
  thermal imaging.
\newblock {\em Infrared Physics \& Technology}, 83:182--187, 2017.

\bibitem{kaplani2012detection}
E.~Kaplani.
\newblock Detection of degradation effects in field-aged c-si solar cells
  through ir thermography and digital image processing.
\newblock {\em International Journal of Photoenergy}, 2012.

\bibitem{kim2016automatic}
D.~Kim, J.~Youn, and C.~Kim.
\newblock Automatic detection of malfunctioning photovoltaic modules using
  unmanned aerial vehicle thermal infrared images.
\newblock {\em Journal of the Korean Society of Surveying, Geodesy,
  Photogrammetry and Cartography}, 34(6):619--627, 2016.

\bibitem{kim2016automaticarea}
D.~Kim, J.~Youn, and C.~Kim.
\newblock Automatic photovoltaic panel area extraction from uav thermal
  infrared images.
\newblock {\em Journal of the Korean Society of Surveying, Geodesy,
  Photogrammetry and Cartography}, 34(6):559--568, 2016.

\bibitem{moreton2014dealing}
R.~Moret{\'o}n, E.~Lorenzo, J.Leloux, and J.M. Carrillo.
\newblock Dealing in practice with hot-spots.
\newblock {\em arXiv preprint arXiv:1411.0621}, 2014.

\bibitem{lumareference}
R.~M.~H. Nguyen and M.~S. Brown.
\newblock Why you should forget luminance conversion and do something better.
\newblock In {\em 2017 IEEE Conference on Computer Vision and Pattern
  Recognition}, pages 5920--5928, July 2017.

\bibitem{otsu1979threshold}
N.~Otsu.
\newblock A threshold selection method from gray-level histograms.
\newblock {\em IEEE transactions on systems, man, and cybernetics},
  9(1):62--66, 1979.

\bibitem{Pierdicca_2018}
R.~Pierdicca, E.~Malinverni, F.~Piccinini, M.~Paolanti, A.~Felicetti, and
  P.~Zingaretti.
\newblock Deep convolutional neural network for automatic detection of damaged
  photovoltaic cells.
\newblock {\em International Archives of the Photogrammetry, Remote Sensing and
  Spatial Information Sciences}, XLII-2:893--900, 05 2018.

\bibitem{belleza1}
P.B. Quater, F.~Grimaccia, S.~Leva, M.~Mussetta, and M.~Aghaei.
\newblock Light unmanned aerial vehicles (uavs) for cooperative inspection of
  pv plants.
\newblock {\em Photovoltaics, IEEE Journal of}, 4:1107--1113, 07 2014.

\bibitem{reza2004realization}
A.M. Reza.
\newblock Realization of the contrast limited adaptive histogram equalization
  (clahe) for real-time image enhancement.
\newblock {\em Journal of VLSI signal processing systems for signal, image and
  video technology}, 38(1):35--44, 2004.

\bibitem{roerdink2000watershed}
J.B. Roerdink and A.~Meijster.
\newblock The watershed transform: Definitions, algorithms and parallelization
  strategies.
\newblock {\em Fundamenta informaticae}, 41(1, 2):187--228, 2000.

\bibitem{tsanakas2013detection}
J.A. Tsanakas and P.N Botsaris.
\newblock On the detection of hot spots in operating photovoltaic arrays
  through thermal image analysis and a simulation model.
\newblock {\em Materials Evaluation}, 71(4), 2013.

\bibitem{tsanakas2015fault2}
J.A. Tsanakas, D.~Chrysostomou, P.N. Botsaris, and A.~Gasteratos.
\newblock Fault diagnosis of photovoltaic modules through image processing and
  canny edge detection on field thermographic measurements.
\newblock {\em International Journal of Sustainable Energy}, 34(6):351--372,
  2015.

\bibitem{tsanakas2015fault}
J.A. Tsanakas, G.~Vannier, A.~Plissonnier, D.L Ha, and F.~Barruel.
\newblock Fault diagnosis and classification of large-scale photovoltaic plants
  through aerial orthophoto thermal mapping.
\newblock In {\em Proceedings of the 31st European Photovoltaic Solar Energy
  Conference and Exhibition 2015}, pages 1783--1788, 2015.

\bibitem{Xie_2020}
X.~Xiaoping, W.~Xiangui, W.~Xingyu, G.~Xincheng, L.~Ju, and C.~Cheng.
\newblock Photovoltaic panel anomaly detection system based on unmanned aerial
  vehicle platform.
\newblock {\em {IOP} Conference Series: Materials Science and Engineering},
  768:072061, mar 2020.

\end{thebibliography}
\bibliographystyle{plain}

% that's all folks
}
\end{document}